\documentclass[10pt,twocolumn,letterpaper]{article}

\usepackage{cvpr}              

\usepackage[dvipsnames]{xcolor}

\definecolor{cvprblue}{rgb}{0.21,0.49,0.74}
\usepackage[pagebackref,breaklinks,colorlinks,allcolors=cvprblue]{hyperref}
\usepackage{marvosym}

\usepackage[utf8]{inputenc} 
\usepackage[T1]{fontenc}    
\usepackage{amsfonts}       
\usepackage{nicefrac}       
\usepackage{microtype}      
\usepackage{dsfont}
\usepackage{mathrsfs}
\usepackage{multirow}
\usepackage[ruled,linesnumbered]{algorithm2e}
\usepackage{colortbl}
\usepackage{pifont}
\usepackage{siunitx}
\usepackage{listings}
\usepackage{newtxtt}
\usepackage{sourcesanspro}

\usepackage{efbox}
\efboxsetup{linecolor=blue,linewidth=0.5pt}

\usepackage{bm}

\usepackage{etoc}

\usepackage{float}

\usepackage{arydshln}
\input{include/macros}

\def\paperID{186} 
\def\confName{CVPR}
\def\confYear{2026}

\title{\alg: A Feature Distillation Enhanced Multi-Expert Ensemble Framework for Robust AI-generated Image Detection}

\newcommand*{\affaddr}[1]{#1} %
\newcommand*{\affmark}[1][]{\textsuperscript{#1}}
\author{%
Zhilin Tu\affmark[1] \quad  Kemou Li\affmark[2] \quad Fengpeng Li\affmark[3]   \quad Jianwei Fei\affmark[4] \quad Jiamin Zhang\affmark[2] \quad Haiwei Wu\affmark[1]\textsuperscript{\Letter}\vspace{1.5mm}\\
\affaddr{\normalsize \affmark[1] School of Computer Science and Engineering, University of Electronic Science and Technology of China} \\
\affaddr{\normalsize \affmark[2] State Key Laboratory of Internet of Things for Smart City, University of Macau} \\
\affaddr{\normalsize \affmark[3] PRADA Lab, King Abdullah University of Science and Technology}\\
\affaddr{\normalsize \affmark[3] Department of Information Engineering, University of Florence}
}

\begin{document}

\etocdepthtag.toc{mtchapter}

\maketitle

\begin{abstract}
The rapid iteration and widespread dissemination of deepfake technology have posed severe challenges to information security, making robust and generalizable detection of AI-generated forged images increasingly important. In this paper, we propose \alg, an AI-generated image detection framework that integrates feature distillation with a multi-expert ensemble, developed for the NTIRE Challenge on Robust AI-Generated Image Detection in the Wild. The framework explicitly targets three practical bottlenecks in real-world forensics: degradation interference, insufficient feature representation, and limited generalization.

Concretely, we build a four-backbone Vision Transformer (ViT) ensemble composed of CLIP and SigLIP variants to capture complementary forensic cues. To improve data coverage, we expand the training set and introduce comprehensive degradation modeling, which exposes the detector to diverse quality variations and synthesis artifacts commonly encountered in unconstrained scenarios. We further adopt a two-stage training paradigm: the model is first optimized with a standard binary classification objective, then refined by dense feature-level self-distillation for representation alignment. This design effectively mitigates overfitting and enhances semantic consistency of learned features.

At inference time, the final prediction is obtained by averaging the probabilities from four independently trained experts, yielding stable and reliable decisions across unseen generators and complex degradations. Despite the ensemble design, the framework remains efficient, requiring only about 10 GB peak GPU memory. Extensive evaluations in the NTIRE challenge setting demonstrate that \alg achieves strong robustness and generalization under diverse ``in-the-wild'' conditions, offering an effective and practical solution for real-world deepfake image detection.

\end{abstract}

\newcommand{\authsym}[1]{\makebox[1em][r]{\textsuperscript{#1}}}

\renewcommand{\thefootnote}{\fnsymbol{footnote}}
\footnotetext{\authsym{\Letter}Corresponding author: Haiwei Wu (haiweiwu@uestc.edu.cn).}

\section{Introduction}
GenAI, represented by GANs~\cite{goodfellow2014generative} and diffusion models~\cite{ho2020denoising}, has advanced rapidly. While boosting innovation in entertainment, malicious AI-generated content (AIGC) can trigger cognitive mislead and trust crisis~\cite{li2026aegis,li2026llm}. To address this problem, numerous forensics~\cite{easyspot2020, wu2023generalizable, wong2025adcd, wu2025rethinking, luo2024lare, kong2025moe, kong2025pixel, cui2024survey,wong2025fontguard}, especially those based on zero-/few-shot paradigms, have been proposed in recent years to better defend against the emerging new GenAI.

Currently, generalizable AIGC detection methods increasingly aim to reduce dependence on generator-specific supervision by learning transferable forensic representations across synthesis families. Representative directions include universal detectors trained for cross-model transfer~\cite{unidet}, zero-shot/open-set forensic formulations~\cite{zed}, and representation learning strategies grounded in large pre-trained vision--language models such as CLIP~\cite{radford2021clip}. In parallel, diffusion-era detection has shifted toward process-aware evidence, where reconstruction and denoising dynamics are explicitly exploited to infer authenticity~\cite{ho2020denoising,dire,aero,luo2024lare}. These developments indicate a clear transition from handcrafted fingerprint cues to mechanism-aware and representation-centric forensics.

\begin{figure*}[t]
    \centering
    \includegraphics[width=0.8\textwidth]{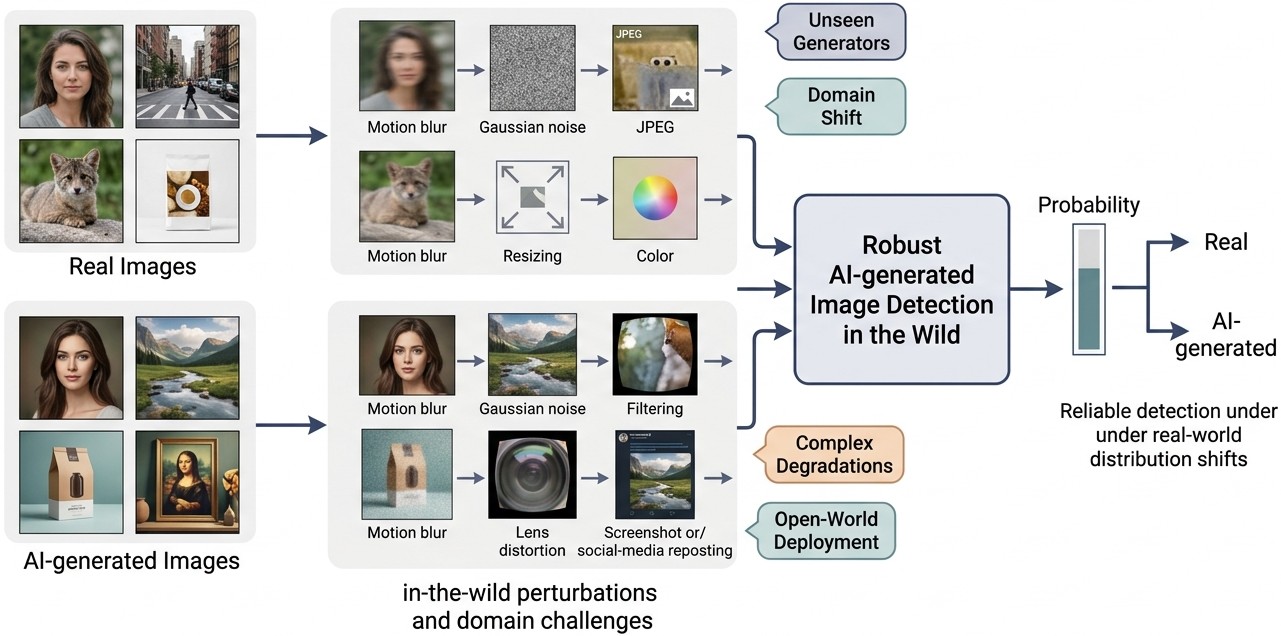}
    \caption{\textbf{Illustration of the robust AI-generated image detection task in the wild.} Real and AI-generated images are both exposed to complex perturbations encountered in practical media pipelines, including motion blur, Gaussian noise, JPEG compression, resizing, color adjustment, lens distortion, filtering, and screenshot/social-media reposting effects. These factors induce severe domain shifts and interactions with unseen generators. A practical detector should therefore produce calibrated probabilities and robust real/AI predictions under open-world deployment constraints.}
    \label{fig:intro_task_illustration}
\end{figure*}

To provide an intuitive overview of our target problem, Fig.~\ref{fig:intro_task_illustration} summarizes the robust AI-generated image detection task in realistic deployment. In practice, both real and AI-generated images may be corrupted by diverse in-the-wild perturbations (\emph{e.g.}, blur, noise, compression, resizing, color distortion, lens distortion, and social-media reposting artifacts), while detectors must remain reliable under unseen generators and domain shifts. The objective is therefore to output stable probabilities and accurate real-vs.-AI decisions in open-world scenarios.

Despite this progress, existing methods still face three critical limitations in real-world deployment. First, domain and distribution shifts remain severe: detectors trained on a fixed generator pool or curated benchmarks can degrade substantially when encountering new content domains, post-processing pipelines, or unseen generators~\cite{genimage,unidet}. Second, prediction stability under image corruption and perturbation is still insufficient, which is problematic in practical media channels where compression, blur, and transmission noise are ubiquitous~\cite{hendrycks2019benchmarking,hendrycks2020augmix}. Third, open-set and zero-shot settings are intrinsically sensitive to reference priors and decision calibration, often leading to unstable false-positive/false-negative trade-offs under challenging ``in-the-wild'' conditions~\cite{zed,unidet}.

From a deployment perspective, a practical detector should satisfy three requirements simultaneously: (1) stable generalization under degradation and distribution shift, (2) consistent representation learning that is less sensitive to superficial artifacts, and (3) efficient inference for large-scale screening. This observation motivates us to move beyond single-model optimization and design a unified framework that improves robustness at the data, feature, and decision levels in a coordinated manner.

To address the above limitations, we propose a robust detection framework, named \textbf{\alg}, which shifts the focus from plain binary classification toward learning degradation-invariant and semantically consistent representations. Specifically, we combine a heterogeneous CLIP--SigLIP multi-expert ensemble with extensive data expansion and a two-stage distillation paradigm that enforces structured feature alignment. The resulting system explicitly couples complementary backbone priors, richer in-the-wild data exposure, and dense representation regularization, thereby improving robustness against unseen generators and complex corruption patterns while maintaining practical efficiency.

Contributions of this work can be summarized as follows:

\begin{enumerate}
    \item We design a high-performance heterogeneous ensemble that integrates CLIP ViT-L/14 and SigLIP So400M experts, leveraging complementary pre-training characteristics to improve robustness and discrimination while preserving an efficient inference footprint of approximately 10 GB.
    \item We improve resilience to in-the-wild distribution shift through joint data expansion and degradation modeling, incorporating state-of-the-art diffusion/adversarial samples and diversified corruption operators (blur, noise, compression, and digital distortions) to better approximate real deployment conditions.
    \item We propose a two-stage optimization strategy with dense feature-level self-distillation, which regularizes representation learning beyond vanilla classification, mitigates overfitting to limited benchmark distributions, and yields more consistent feature structures across unseen generation methods.
\end{enumerate}

\textbf{Organization.}
The rest of this paper is organized as follows.
Section~\ref{sec:related-works} reviews related works on AI-generated image detection. 
Section~\ref{sec:data} introduces the experimental data used in the NTIRE 2026 Robust AI-Generated Image Detection in the Wild competition.
Section~\ref{sec:method} describes our proposed \alg, including training and inference phases. 
Section~\ref{sec:experiments} reports experimental results and ablations. 
Finally, Section~\ref{sec:conclusion} concludes the paper.

\section{Related Works}
\label{sec:related-works}

\subsection{Deepfake and Image Forgery Detection}

Early deepfake and image-forgery research established the core benchmarks and forensic paradigms for manipulation detection. FaceForensics++~\cite{faceforensics2019} and the DFDC dataset~\cite{dolhansky2020dfdc} significantly advanced large-scale evaluation by providing diverse manipulated samples and realistic compression settings. Meanwhile, early generalized detectors showed that synthetic imagery contains exploitable spectral and upsampling artifacts. Representative studies demonstrated strong discriminative signals in both spatial and frequency domains, including CNN artifact analysis and Fourier-spectrum inconsistencies~\cite{wang2020cnnspot,durall2020watch,dzanic2020fourier}. Recent work further extends forensic cues to robust font watermark consistency and editing-fingerprint learning with self-augmentation~\cite{wong2025fontguard,wu2026editprint}. These works laid the foundation for moving from identity-specific deepfake detection toward broader synthetic-image forensics.

\subsection{Detection in the Diffusion Era}

With diffusion models becoming dominant generative engines~\cite{ho2020denoising}, detection methods have increasingly focused on tracing generation-process traces rather than static texture fingerprints. Process-aware approaches exploit reconstruction dynamics, latent inversion errors, or model-specific residual behaviors to distinguish real from generated images. Notable examples include DIRE~\cite{dire}, AEROBLADE~\cite{aero}, and LaRE$^2$~\cite{luo2024lare}, which improve robustness against high-fidelity diffusion outputs. Compared with earlier GAN-oriented cues, these methods better capture intrinsic synthesis mechanisms and therefore offer stronger transferability to unseen diffusion variants.

\subsection{Generalization and Open-World Robustness}

Generalization across generators and domains is now a central objective in AIGC forensics. Universal detection paradigms explicitly optimize cross-generator transfer~\cite{unidet}, while large-scale benchmarks such as GenImage~\cite{genimage} reveal the practical difficulty of maintaining performance under unseen generator families and real-world distortions. In parallel, zero-shot formulations~\cite{zed} aim to reduce annotation dependence, but robustness remains closely tied to distribution shift handling and calibration quality. Robust-learning research on common corruptions and augmentation-based regularization further suggests that corruption-aware training is essential for stable deployment~\cite{hendrycks2019benchmarking,hendrycks2020augmix}.
Complementary studies on label-noise robust optimization also support stable supervision under noisy or shifted data~\cite{rml2024,DBLP:journals/ijcv/LiLWHTZ25}

\subsection{Pre-trained Backbones and Distillation}

Recent progress in large-scale pre-training provides strong transferable priors for forensic representation learning. Vision Transformer architectures~\cite{dosovitskiy2021vit}, CLIP-style contrastive pre-training~\cite{radford2021clip}, and SigLIP's sigmoid-loss objective~\cite{zhai2023siglip} have each demonstrated strong cross-domain representation capabilities. Complementarily, knowledge distillation and self-distillation paradigms~\cite{hinton2015distilling,zhang2019byot} offer effective regularization for improving feature consistency and reducing overfitting. Our framework is aligned with this line of work by combining heterogeneous pre-trained experts with dense feature-level self-distillation to enhance robustness under real-world degradations and generator shifts.

\section{Competition Data}
\label{sec:data}

The experimental data for this work is sourced from the NTIRE 2026 Robust AI-Generated Image Detection in the Wild competition hosted on Codabench (Competition ID: 12761), which is meticulously constructed to simulate the complex and unconstrained real-world scenarios of AI-generated image forensics. The dataset is stratified into multiple functional splits with progressive difficulty, including a toy dataset, training set, two validation subsets (1st Validation and Hard Validation), and two test subsets (Public Test and Private Test), with the total scale of labeled training data reaching approximately 277,000 images. Each split is composed of a balanced mix of authentic natural images and AI-generated forged images, where the forged samples are synthesized by an expanding set of state-of-the-art generator models across splits—ranging from 10 generators for the toy dataset to 35 advanced generators for the Private Test set—covering the latest progress in AI image generation and ensuring the dataset’s representativeness of real-world forgery diversity.
For clarity, Table~\ref{tab:competition_data_summary} summarizes the key protocol settings of each competition split.

\begin{table*}[t]
    \centering
    \small
    \setlength{\tabcolsep}{5pt}
    \caption{Summary of NTIRE 2026 robust AI-generated image detection competition data splits.}
    \label{tab:competition_data_summary}
    \begin{tabular}{lcccll}
        \toprule
        Split & \# Generators & \# Distortions & Distortion Profile & Labels & Role \\
        \midrule
        Toy & 10 & 0 & Clean only & Yes & Debug/sanity check \\
        Training & 20 & 0 & Clean only & Yes & Supervised training \\
        1st Validation & 25 & 5 & Unseen simple distortions & No & Model selection \\
        Hard Validation & 25 & 5 & Harder simple distortions & No & Robust validation \\
        Public Test & 30 & 7 & Unknown mixed distortions & No & Public leaderboard \\
        Private Test & 35 & 9 & Unknown mixed, hardest setting & No & Final ranking \\
        \bottomrule
    \end{tabular}
\end{table*}

A core design highlight of this dataset is its comprehensive and tiered degradation modeling implemented via the official distortion pipeline, which is specifically tailored to mimic the various image quality degradations that both real and AI-generated images encounter during real-world acquisition, transmission, and post-processing. For the training set, the raw authentic and AI-generated images are provided without additional artificial transformations, serving as the baseline data for model initialization and basic feature learning. In contrast, all subsequent validation and test splits integrate multi-dimensional and increasingly complex degradation operations, with the number of transformation types escalating in a stepwise manner: the 1st Validation and Hard Validation sets apply 5 types of simple distortions, the Public Test set combines 7 types of simple and complex degradations, and the Private Test set further extends to 9 types of mixed distortions. These official degradation operations encompass a full spectrum of real-world image corruptions, including standard blurring effects, multi-type noise patterns, lossy compression artifacts, as well as advanced optical distortions (e.g., lens aberration, light refraction) and digital post-processing distortions (e.g., color adjustment, resizing, filtering), which are consistent with the actual quality variations of images spread on social media, e-commerce platforms, and other internet channels.

To further challenge the robustness and generalization of detection models and align with the practical demands of real-world forensics (where unseen image distortions and forgery methods are ubiquitous), the competition dataset introduces unknown degradation transformations and novel generator models in the validation and test phases that are not present in the training set. The 1st Validation set expands the generator models from 20 (training set) to 25 and adds unseen simple distortions, while the Hard Validation set, as a closer proxy to the test data, retains the 25 generator models and 5 simple distortions but optimizes the distortion intensity and combination to increase detection difficulty. The Public Test set further scales up to 30 generator models and 7 mixed simple/complex unknown degradations, and the Private Test set— the most challenging split—employs 35 cutting-edge and unseen AI generator models alongside 9 sophisticated mixed distortion types, with manual verification and fairness checks applied to ensure the validity of the test results. Notably, no labels (including real/AI-generated labels and clean/distorted labels) are provided for all validation and test splits, which requires the detection model to autonomously learn degradation-invariant discriminative features and avoid overfitting to specific distortion patterns or generator artifacts.



\begin{figure*}[t!]
	\centering
		\includegraphics[width = \textwidth]{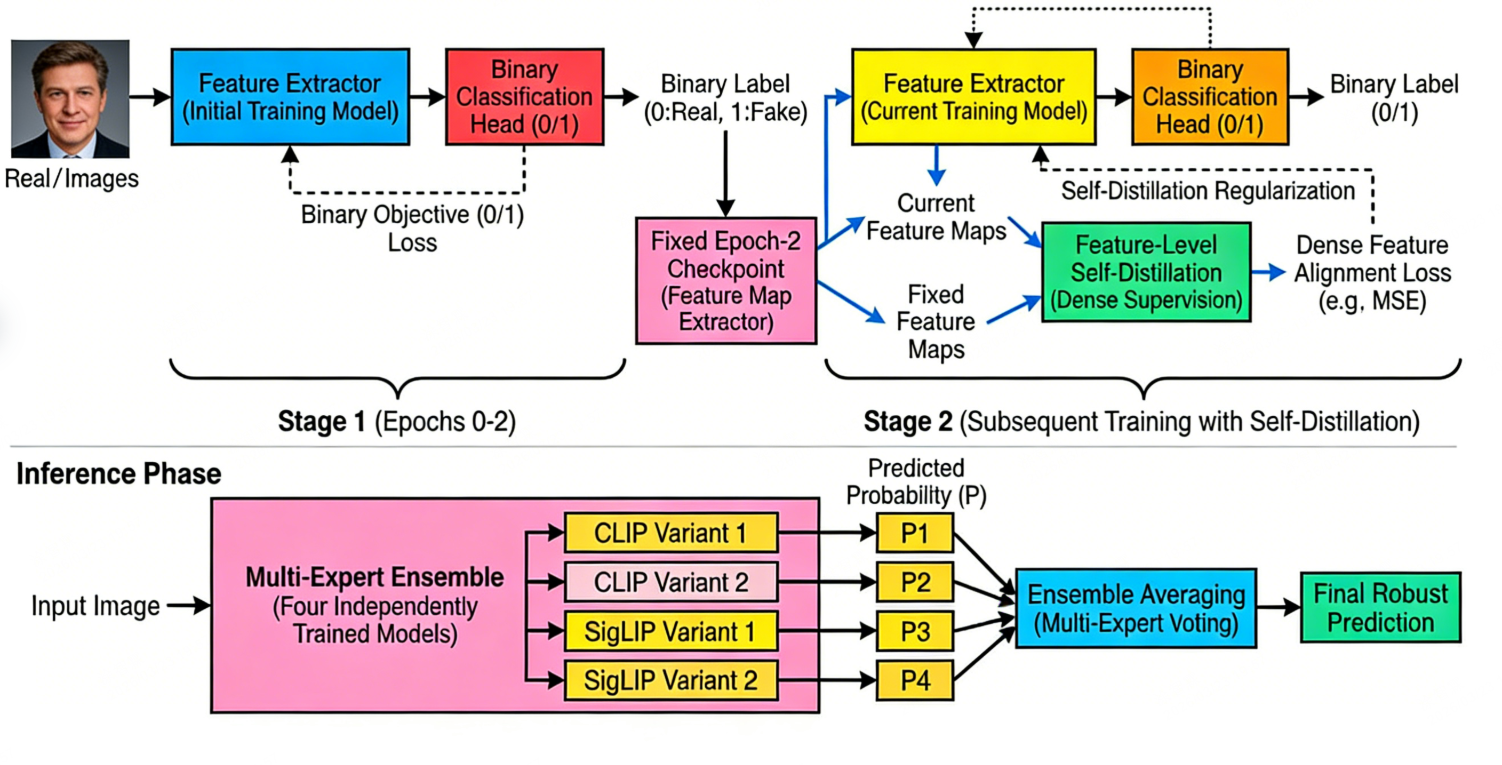}
	\caption{\textbf{Overview of the proposed \alg framework}. Our approach utilizes a multi-expert ensemble consisting of four high-capacity backbones. The training follows a two-stage paradigm: Stage 1 establishes a discriminative baseline, while Stage 2 employs dense feature-level self-distillation to enhance structural consistency. Robustness is further ensured through multi-source data expansion and a comprehensive 35-algorithm degradation library.}
	\label{fig:framework}
\end{figure*}

\section{Proposed Method: \alg}
\label{sec:method}

As illustrated in Figure~\ref{fig:framework}, the \alg framework is designed to address the challenges of detecting high-fidelity AI-generated images under complex real-world conditions. Our method harmonizes diverse Vision-Language pre-trained backbones with a robust distillation-based training paradigm and extensive data augmentation.

\subsection{Architecture Overview}
The core of our framework is a multi-expert ensemble comprising four Vision Transformer (ViT) backbones: two \textbf{CLIP ViT-L/14} models and two \textbf{SigLIP So400M} models. On top of each backbone, a lightweight classification head is attached to map high-dimensional features to binary predictions. To accommodate architectural requirements, input images are preprocessed with backbone-specific resizing and cropping ($224 \times 224$ for CLIP and $384 \times 384$ for SigLIP). Despite the multi-model design, the ensemble remains efficient, maintaining a peak GPU memory footprint of approximately 10 GB during inference.

\subsection{Data Strategy and Degradation Modeling}
To bridge the gap between laboratory samples and in-the-wild forgeries, we implement a two-pronged data strategy:

\textbf{Multi-source Data Expansion.} We incorporate approximately 205,000 external samples to improve generalization. This includes \textit{DiTFake} (30k) for capturing artifacts of modern Diffusion Transformers (Flux, SD3), \textit{DiffFace} (70k) for localized facial edit detection, and \textit{De-Factify} (42k) combined with \textit{Deepfake-60K} (60k) to handle social media noise and traditional GAN forgeries.

\textbf{Comprehensive Degradation Library.} We develop an extensive augmentation library, \texttt{distortions\_extended.py}, featuring 35 specific algorithms across eight categories: blurs (motion, atmospheric), advanced sensor noise (Poisson, ISO), compression artifacts (JPEG 2000, ringing), color distortions, geometric transforms, environmental effects (fog, rain, snow), sensor blooming, and random occlusions. During training, we alternate equally between the official pipeline and this extended library to force the model to focus on intrinsic forgery signatures rather than low-level quality fluctuations.

\subsection{Two-Stage Training with Dense Supervision}
We propose a two-stage training strategy to refine feature representations:
\begin{itemize}
    \item \textbf{Stage 1: Fundamental Learning.} The model is optimized using standard binary cross-entropy loss to learn the basic decision boundary between real and forged distributions.
    \item \textbf{Stage 2: Dense Feature Distillation.} We freeze the Stage 1 checkpoint as a reference. The active model then undergoes self-distillation, where its intermediate feature maps are aligned with reference features via dense supervision. This paradigm, combined with the classification objective, ensures structural and semantic consistency while mitigating overfitting.
\end{itemize}

\subsection{Multi-Expert Ensemble Inference}
During inference, we aggregate the predicted probabilities from the four independently trained expert variants through simple averaging. This ensemble strategy effectively reduces prediction variance and improves stability under significant distribution shifts, providing a robust solution for deepfake detection in real-world scenarios.

\subsection{Backbone Perspective}


To identify an optimal architectural foundation, we conduct a large-scale empirical evaluation over a diverse set of feature extractors $\mathcal{F} = \{f_1, f_2, \dots, f_n\}$, including CLIP, SigLIP, MoCo, Swin Transformer, ConvNeXt, DINO, CoCa, and BEiT. We evaluate each extractor based on representation quality, computational efficiency, and robustness to synthetic artifacts. Empirically, we observe that vision--language pre-trained models—specifically CLIP ($f_{\text{CLIP}}$) and SigLIP ($f_{\text{SigLIP}}$)—consistently achieve the superior trade-off between discriminative capability and cross-generator generalization. Our final framework utilizes a multi-expert ensemble of four backbones: two CLIP ViT-L/14 models and two SigLIP So400M variants. Input images are preprocessed with backbone-specific transformations $\mathcal{T}_f$, where $x \in \mathbb{R}^{224\times224}$ for CLIP and $x \in \mathbb{R}^{384\times384}$ for SigLIP.

\subsection{Data Perspective}


We enhance model robustness through two complementary strategies: training set expansion and comprehensive degradation modeling.
Our augmentation policy is additionally inspired by frequency-domain generative amplitude mix-up, which improves robustness under perturbations~\cite{li2024dat}.
First, we expand the training distribution $\mathcal{D}_{\text{train}}$ by integrating images generated from a suite of contemporary synthesis techniques $\mathcal{S}$, including state-of-the-art diffusion models, reconstruction pipelines, and domain-transfer methods. This expansion also includes ``hard-negative'' samples generated via targeted adversarial attacks to expose the detector to emerging forgery artifacts.
Second, to simulate ``in-the-wild'' quality variations, we define a degradation function $\phi(x)$ that integrates an extended array of operators, including complex blurs, noise patterns, and advanced digital distortions. By training on these perturbed samples, we force the model to learn a degradation-invariant representation $z = f(\phi(x))$, significantly improving resilience to distribution shifts caused by varying imaging conditions or transmission errors.

\subsection{Training Perspective}


We propose a two-stage training paradigm to capture richer artifact representations beyond vanilla binary classification.
In Stage 1 (Epochs 0-2), the model $f_\theta$ is trained using a standard binary cross-entropy objective $\mathcal{L}_{\text{BCE}}$ based on labels $y \in \{0, 1\}$:
\begin{equation}
    \mathcal{L}_{\text{stage1}} = -\frac{1}{N} \sum_{i=1}^{N} [y_i \log(\hat{y}_i) + (1-y_i) \log(1-\hat{y}_i)].
\end{equation}

In Stage 2, we introduce a feature-level self-distillation mechanism. We utilize a fixed checkpoint from epoch 2 as a teacher model to extract dense feature maps $M_{\text{fixed}}$. The current training model is guided to align its feature maps $M_{\text{current}}$ with these distilled targets through a dense alignment loss:
\begin{equation}
    \mathcal{L}_{\text{distill}} = \| M_{\text{current}} - M_{\text{fixed}} \|^2_2.
\end{equation}
This dense supervision acts as a regularizer, ensuring semantic consistency and mitigating overfitting to the competition dataset.
We further emphasize preserving stable core cues during optimization, consistent with core feature-aware robust training principles~\cite{DBLP:journals/tifs/LiLWTZ25}. 
During the Inference Phase, the final prediction $P_{\text{final}}$ is derived by averaging the probabilities from the multi-expert ensemble:
\begin{equation}
    P_{\text{final}} = \frac{1}{4} \sum_{k=1}^{4} P_k,
\end{equation}
where $P_k$ represents the output of the $k$-th independently trained CLIP or SigLIP variant.

\subsection{Inference Perspective}
The inference phase of our proposed framework is engineered to prioritize decision robustness and architectural synergy, moving beyond the limitations of single-model predictions. As illustrated in our architectural pipeline, the final classification is determined through a sophisticated Multi-Expert Ensemble strategy comprising four independently trained deep learning backbones: two CLIP ViT-L/14 variants and two SigLIP So400M variants. This heterogeneous composition is strategically chosen to capture a diverse spectrum of forgery artifacts, as the distinct pre-training objectives of CLIP and SigLIP provide complementary perspectives on image-text alignment and visual representation.

During the deployment stage, an input image $x_{test}$ undergoes backbone-specific preprocessing $\mathcal{T}_k$, ensuring that the spatial resolution and normalization parameters are strictly aligned with the requirements of each expert model. Specifically, the CLIP-based experts process inputs at a standard $224 \times 224$ resolution, while the SigLIP-based experts utilize a higher $384 \times 384$ resolution to capture finer-grained structural anomalies. Each expert $f_k$ extracts high-dimensional feature representations, which are subsequently mapped to a binary logit by a lightweight classification head.

The core of our inference reliability lies in the Multi-Expert Voting mechanism. Rather than relying on a hard-label consensus, we employ a soft-voting strategy by averaging the predicted probabilities $P_k$ from all four models. Formally, the final robust prediction $P_{final}$ is computed as:
\begin{equation}
    P_{final} = \frac{1}{K} \sum_{k=1}^{K} P_k(y=1| \mathcal{T}_k(x_{test}); \theta_k),
\end{equation}
where $K=4$ represents the total number of experts and $\theta_k$ denotes the parameters of the $k$-th model optimized through our two-stage self-distillation paradigm. This averaging process effectively dampens the influence of individual model biases and reduces the variance of the detector, leading to more stable performance across unseen generation methods.

Despite the ensemble's structural complexity, the system maintains high operational efficiency. By optimizing the execution of the four Transformers, the framework achieves a peak GPU memory footprint of approximately 10 GB, making it feasible for deployment on standard consumer-grade hardware or cloud-based screening APIs. This balance of high-fidelity detection and computational feasibility ensures that our model can effectively handle the diverse real-world distributions and varying imaging conditions typical of ``in the wild'' AIGC content.

\begin{table*}[t]
\centering
\caption{Quantitative comparison of different backbones and strategies on the NTIRE Challenge. ``Dense Labels'' indicates the application of our Stage 2 feature distillation. The best results for each set are bolded.}
\label{tab:comparison}
\resizebox{\textwidth}{!}{%
\begin{tabular}{@{}lcccccc@{}}
\toprule
& & & & \multicolumn{1}{c}{Online Validation} & \multicolumn{1}{c}{Online Test} \\ \cmidrule(lr){5-5} \cmidrule(lr){6-6}
Backbone & Architecture & Parameters & Dense Labels & Robust ROC AUC & Robust Hard ROC AUC \\ \midrule
Swin-T & Transformer & 28M & No & 0.785 & N/A \\
CLIP-RN50x64 & CNN & 420M & No & 0.828 & N/A \\
MOCOV3 & Transformer & 86M & No & 0.851 & N/A \\
CONVNEXT & CNN & 89M & No & 0.855 & N/A \\
BEIT & Transformer & 86M & No & 0.857 & N/A \\
DINOV2 & Transformer & 86M & No & 0.868 & N/A \\
CLIP-L/14 & Transformer & 304M & No & 0.8926 & N/A \\ \midrule
SigLIP-400M & Transformer & 400M & Yes & 0.926 & N/A \\
CLIP-L/14 & Transformer & 304M & Yes & \textbf{0.934} & N/A \\ \midrule
SigLIP-400M & Transformer & 400M & Yes & N/A & 0.848 \\
CLIP-L/14 & Transformer & 304M & Yes & N/A & 0.845 \\
\textbf{2 CLIP-L/14 + 2 SigLIP-400M} & \textbf{Ensemble} & \textbf{1400M} & \textbf{Yes} & \textbf{N/A} & \textbf{0.856} \\ \bottomrule
\end{tabular}%
}
\end{table*}

\section{Experiments}
\label{sec:experiments}

\subsection{Implementation Details}
\textbf{Hardware and Training Environment.} 
Our framework is implemented using PyTorch and trained on a high-performance cluster equipped with two NVIDIA H100 (80GB) GPUs. To balance the high computational requirements of Large-scale Vision Transformer backbones (CLIP ViT-L/14 and SigLIP So400M) with a substantial effective batch size, we employ \textbf{Mixed Precision (FP16)} training. Under a global batch size of 128, the peak VRAM consumption is approximately \textbf{60 GB per GPU}, ensuring efficient throughput and stable gradient updates.

\textbf{Two-Stage Training Strategy.} 
The optimization process is divided into two distinct phases:
\begin{itemize}
    \item \textbf{Stage 1 (Initial Warm-up):} We first optimize the backbone and linear classification heads using a standard Binary Cross-Entropy (BCE) loss for 2 epochs. This stage establishes a robust baseline discriminative capability and initializes the ``Teacher'' model with structured forensic knowledge.
    \item \textbf{Stage 2 (Contrastive Refinement):} We introduce the \textbf{Contrastive Representation Distillation (CRD)} loss. By maintaining a MoCo-style negative buffer, we explicitly pull representations of the same image (under different augmentations) together while pushing different images apart in the embedding space. The joint objective is defined as: $\mathcal{L}_{total} = \mathcal{L}_{BCE} + \lambda \mathcal{L}_{CRD}$.
\end{itemize}

\textbf{Dynamic Momentum Teacher.} 
Instead of a static Exponential Moving Average (EMA), we argue that the teacher should adaptively evolve. We implement a \textbf{Cosine-scheduled Momentum Update} for the teacher's weights:
\begin{equation}
m = m_{max} - (m_{max} - m_{base}) \cdot \frac{\cos(\pi \cdot \frac{step_{global}}{step_{total}}) + 1}{2}
\end{equation}
where we set $m_{base} = 0.99$ and $m_{max} = 0.9999$. This schedule allows the teacher to remain plastic during the early refinement phase and reach maximum stability toward the end of training, effectively preventing representation drift in unconstrained scenarios.

\subsection{Performance Analysis}

\textbf{Quantitative Results.} 
We conduct extensive experiments to evaluate the effectiveness of different backbones and training strategies. As summarized in Table~\ref{tab:comparison}, we observe that Vision-Language Pre-trained (VLP) models, such as CLIP and SigLIP, significantly outperform standard vision backbones like Swin-T and ConvNeXt. Specifically, the CLIP-L/14 backbone achieves a Robust ROC AUC of 0.8926 even without dense supervision.

A key finding is the efficacy of our proposed \textbf{two-stage strategy (Dense Labels)}. By incorporating dense feature-level distillation, the performance of CLIP-L/14 further improves from 0.8926 to \textbf{0.934} on the Online Validation set. Finally, our multi-expert ensemble, which integrates two CLIP-L/14 and two SigLIP-400M models, achieves the best generalization on the most challenging \textbf{Online Test (Hard)} set with a Robust AUC of \textbf{0.856}, demonstrating its superior robustness in-the-wild.

\textbf{Impact of External Data.}
In addition to backbone selection and dense supervision, we systematically investigate the contribution of multi-source external data to model robustness. To improve generalization across unseen generative pipelines, we augment the official training subset with several complementary datasets covering diverse generative mechanisms and real-world distortions.

Specifically, the inclusion of data generated by modern diffusion and Diffusion Transformer (DiT) architectures plays a crucial role. Samples from datasets such as \textit{DiTFake} introduce artifacts produced by recent high-fidelity generative models (e.g., Flux and SD-based pipelines), allowing the model to better adapt to emerging synthesis paradigms. Meanwhile, the integration of localized manipulation samples from \textit{DiffFace} enhances sensitivity to subtle regional edits, especially around semantically critical facial regions such as eyes and nose. These samples are particularly challenging because the overall image remains visually realistic while only local structures are modified.

We further incorporate real-world noisy imagery from datasets such as \textit{De-Factify}, which introduces compression artifacts, overlays, and meme-style perturbations commonly observed in social media environments. This data helps the model maintain stable performance under complex noise conditions. Additionally, traditional deepfake datasets such as \textit{Deepfake-60K} provide diverse legacy manipulation patterns, ensuring coverage of classical GAN-based synthesis artifacts. Overall, the combination of these heterogeneous data sources significantly expands the diversity of generative distributions observed during training, leading to improved robustness under domain shifts.

\textbf{Impact of Extended Degradation Strategies.}
Beyond data diversity, we also investigate the influence of large-scale degradation modeling on detection robustness. In addition to the official distortion pipeline, we design an extended degradation module that introduces a wide range of realistic image corruptions. During training, distortions from both the official pipeline and our extended distortion library are applied with approximately equal probability, ensuring balanced exposure to multiple degradation patterns.

The extended degradation framework contains eight major categories of image corruptions, including blur, noise, compression artifacts, color distortions, geometric transformations, environmental effects, sensor-level noise, and occlusion-based perturbations. Representative operations include motion blur, defocus blur, JPEG2000 compression artifacts, color casting, perspective warping, fog simulation, sensor blooming, and random occlusion. These degradations simulate a wide spectrum of real-world image formation and transmission processes, effectively increasing the difficulty of the training task.

Empirically, we observe that introducing extended degradations significantly improves robustness on challenging validation scenarios, particularly under heavy compression and environmental noise conditions. Models trained with mixed degradation strategies demonstrate improved stability across unseen data distributions and maintain higher detection confidence under strong image quality degradation. This finding highlights the importance of modeling realistic image corruption patterns for robust synthetic image detection.

\textbf{Overall Analysis.}
Combining dense supervision, multi-source external data, and extended degradation strategies yields a substantial improvement in detection robustness. Each component contributes complementary benefits: dense-label supervision enhances feature discrimination, external datasets increase generative diversity, and extended degradations improve resilience to quality degradation. Together, these design choices enable our final multi-expert framework to achieve strong generalization performance across both validation and hard test environments, demonstrating its effectiveness for real-world synthetic image detection tasks.


\section{Conclusions}\label{sec:conclusion}

In this paper, we presented \alg, a practical framework for robust AI-generated image detection in the wild. Unlike conventional settings that assume relatively clean and closed distributions, our work targets a more realistic deployment scenario in which detectors must remain reliable under severe degradations, domain shifts, and previously unseen generator families. To address these challenges, we combined a heterogeneous CLIP–SigLIP multi-expert ensemble with training data expansion, comprehensive degradation modeling, and a two-stage feature distillation strategy, enabling the detector to learn more stable and semantically consistent forensic representations beyond superficial artifacts.

Extensive experiments on the NTIRE 2026 Robust AI-Generated Image Detection in the Wild challenge validate the effectiveness of our design. The results show that large vision–language pre-trained backbones provide a strong foundation for cross-generator generalization, while feature-level distillation, external data augmentation, and extended degradation strategies each contribute complementary gains in robustness. By integrating these components within a unified framework, our final ensemble achieves strong performance under challenging validation and test conditions, while still maintaining practical inference efficiency for real-world deployment.

Overall, this work highlights that robust synthetic-image forensics should not rely on a single factor alone, but instead requires coordinated improvements at the data, representation, and decision levels. We hope \alg can serve as a strong baseline for real-world AIGC detection and inspire future research on more efficient, better calibrated, and more generalizable forensic systems for open-world scenarios.

{
    \small
    \bibliographystyle{ieeenat_fullname}
    \bibliography{ref}

@String(AAAI = {AAAI})

@article{cui2024survey,
  title={Survey on Fake Information Generation, Dissemination and Detection},
  author={W. Cui and D. Wang and N. Han},
  journal={Chin. J. Electron.},
  volume={33},
  number={3},
  pages={573--583},
  year={2024}
}

@inproceedings{luo2024lare,
  title={LaRE\^{} 2: Latent reconstruction error based method for diffusion-generated image detection},
  author={Y. Luo and J. Du and K. Yan and S. Ding},
  booktitle={Proc. Comput. Vis. Pattern Recogn.},
  pages={17006--17015},
  year={2024}
}

@article{kong2025moe,
  title={MoE-FFD: Mixture of Experts for Generalized and Parameter-Efficient Face Forgery Detection},
  author={C. Kong and A. Luo and P. Bao and Y. Yu and H. Li and Z. Zheng and S. Wang and A. Kot},
  journal={IEEE Trans. Dependable Secure Comput.},
  year={2025},
  pages={1--15},
}

@article{wu2025rethinking,
  title={Rethinking Image Forgery Detection Via Soft Contrastive Learning and Unsupervised Clustering},
  author={H. Wu and Y. Chen and J. Zhou and Y. Li},
  journal={IEEE Trans. Dependable Secure Comput.},
  year={2025},  
  volume={22},
  number={6},
  pages={6296-6308},
}

@inproceedings{wong2025adcd,
  title={ADCD-Net: Robust Document Image Forgery Localization via Adaptive DCT Feature and Hierarchical Content Disentanglement},
  author={K. Wong and J. Zhou and H. Wu and Y. Si and J. Zhou},
  booktitle={Proc. Int. Conf. Comput. Vis.},
  pages={19280--19289},
  year={2025}
}

@article{wu2023generalizable,
  title={Generalizable synthetic image detection via language-guided contrastive learning},
  author={H. Wu and J. Zhou and S. Zhang},
  journal={IEEE Trans. on Artificial Intelligence},
  pages={1--11},
  year={2025}
}

@article{kong2025pixel,
  title={Pixel-Inconsistency Modeling for Image Manipulation Localization},
  author={C. Kong and A. Luo and S. Wang and H. Li and A. Rocha and A. Kot},
  journal={IEEE Trans. Pattern Anal. and Mach. Intell.},
  year={2025},
  volume={47},
  number={6},
  pages={4455--4472},
}

@inproceedings{ho2020denoising,
  title={Denoising diffusion probabilistic models},
  author={J. Ho and A. Jain and P. Abbeel},
  booktitle={Adv. in Neur. Info. Proc. Syst.},
  pages={6840--6851},
  year={2020}
}

@inproceedings{genimage,
  title={Genimage: A million-scale benchmark for detecting ai-generated image},
  author={M. Zhu and H. Chen and Q. Yan and X. Huang and G. Lin and W. Li and Z. Tu and H. Hu and J. Hu and Y. Wang},
  booktitle={Proc. Neural Info. Process. Syst.},
  pages={77771--77782},
  year={2023}
}

@inproceedings{zed,
  title={Zero-shot detection of ai-generated images},
  author={D. Cozzolino and G. Poggi and M. Nie{\ss}ner and L. Verdoliva},
  booktitle={Proc. Eur. Conf. Comput. Vis.},
  pages={54--72},
  year={2024}
}

@inproceedings{aero,
  title={Aeroblade: Training-free detection of latent diffusion images using autoencoder reconstruction error},
  author={J. Ricker and D. Lukovnikov and A. Fischer},
  booktitle={Proc. Comput. Vis. Pattern Recogn.},
  pages={9130--9140},
  year={2024}
}

@inproceedings{dire,
  title={DIRE for Diffusion-Generated Image Detection},
  author={Z. Wang and J. Bao and W. Zhou and W. Wang and H. Hu and H. Chen and H. Li},
  booktitle={Proc. Int. Conf. Comput. Vis.},
  pages={22445--22455},
  year={2023}
}

@inproceedings{unidet,
  title={Towards Universal Fake Image Detectors that Generalize Across Generative Models},
  author={U. Ojha and Y. Li and Y. Lee},
  booktitle={Proc. Comput. Vis. Pattern Recogn.},
  pages={24480--24489},
  year={2023}
}

@inproceedings{convnext,
  title={A convnet for the 2020s},
  author={Z. Liu and H. Mao and C. Wu and C. Feichtenhofer and T. Darrell and S. Xie},
  booktitle={Proc. Comput. Vis. Pattern Recogn.},
  pages={11976--11986},
  year={2022}
}

@inproceedings{moco,
  title={Momentum contrast for unsupervised visual representation learning},
  author={K. He and H. Fan and Y. Wu and S. Xie and R. Girshick},
  booktitle={Proc. Comput. Vis. Pattern Recogn.},
  pages={9729--9738},
  year={2020}
}

@inproceedings{durall2020watch,
  title={Watch your up-convolution: Cnn based generative deep neural networks are failing to reproduce spectral distributions},
  author={R. Durall and M. Keuper and J. Keuper},
  booktitle={Proc. Comput. Vis. Pattern Recogn.},
  pages={7890--7899},
  year={2020}
}

@article{dzanic2020fourier,
  title={Fourier spectrum discrepancies in deep network generated images},
  author={T. Dzanic and K. Shah and F. Witherden},
  journal={Proc. Adv. Neural Inf. Process. Syst.},
  volume={33},
  pages={3022--3032},
  year={2020}
}

@inproceedings{clip,
  title={Learning transferable visual models from natural language supervision},
  author={A. Radford and J. Kim and C. Hallacy and A. Ramesh and G. Goh and S. Agarwal and G. Sastry and A. Askell and P. Mishkin and J. Clark and K. Gretchen and S. Ilya},
  booktitle={Proc. Int. Conf. Mach. Learn.},
  pages={8748--8763},
  year={2021}
}

@inproceedings{rml2024,
  title={Regroup Median Loss for Combating Label Noise},
  author={Li, F. and Li, K. and Tian, J. and Zhou, J.},
  booktitle={Proc. AAAI Conf. Arti. Intell.},
  pages={13474-13482},
  year={2024}
}

@inproceedings{li2024dat,
  title={{DAT}: Improving Adversarial Robustness via Generative Amplitude Mix-up in Frequency Domain},
  author={F. Li and K. Li and H. Wu and J. Tian and J. Zhou},
  booktitle={Proc. Adv. Neural Inf. Process. Syst.},
    pages = {127099--127128},
  year={2024}
}

@article{DBLP:journals/ijcv/LiLWHTZ25,
  author       = {F. Li and
                  K. Li and
                  Q. Wang and
                  B. Han and
                  J. Tian and
                  J. Zhou},
  title        = {{RML++:} Regroup Median Loss for Combating Label Noise},
  journal      = {Int. J. Comput. Vis.},
  volume       = {133},
   pages        = {6400--6421},
number={9},
  year         = {2025}
}

@article{DBLP:journals/tifs/LiLWTZ25,
  author       = {F. Li and
                  K. Li and
                  H. Wu and
                  J. Tian and
                  J. Zhou},
  title        = {Toward Robust Learning via Core Feature-Aware Adversarial Training},
  journal      = {{IEEE} Trans. Inf. Forensics Secur.},
  volume       = {20},
  numbers={9},
  pages        = {6236--6251},
  year         = {2025},
}

@inproceedings{goodfellow2014generative,
  title={Generative adversarial nets},
  author={I. Goodfellow and J. Pouget-Abadie and M. Mirza and B. Xu and D. Warde-Farley and S. Ozair and A. Courville and Y. Bengio},
  booktitle={Proc. Neural Info. Process. Syst.},
  pages={2672--2680},
  year={2014}
}

@inproceedings{easyspot2020,
  title={CNN-generated images are surprisingly easy to spot... for now},
  author={S. Wang and O. Wang and R. Zhang and A. Owens and A. A. Efros},
  booktitle={Proc. Comput. Vis. Pattern Recogn.},
  pages={8695--8704},
  year={2020}
}

@inproceedings{faceforensics2019,
  title={Faceforensics++: learning to detect manipulated facial images},
  author={A. Rossler and D. Cozzolino and L. Verdoliva and C. Riess and J. Thies and M. Nießner},
  booktitle={Proc. Int. Conf. Comput. Vis.},
  pages={1--11},
  year={2019}
}

@inproceedings{li2026aegis,
  title={{AEGIS}: Adversarial Target-Guided Retention-Data-Free Robust Concept Erasure from Diffusion Models},
  author={F. Li and K. Li and Q. Wang and B. Han and J. Zhou},
  booktitle={Proc. Int. Conf. Learn. Representat.},
  year={2026}
}

@article{dolhansky2020dfdc,
  title={The DeepFake Detection Challenge (DFDC) Dataset},
  author={B. Dolhansky and R. Howes and B. Pflaum and N. Baram and C. Canton Ferrer},
  journal={arXiv preprint arXiv:2006.07397},
  year={2020}
}

@inproceedings{wang2020cnnspot,
  title={CNN-Generated Images Are Surprisingly Easy to Spot... For Now},
  author={S.-Y. Wang and O. Wang and R. Zhang and A. Owens and A. A. Efros},
  booktitle={Proc. Comput. Vis. Pattern Recogn.},
  pages={8695--8704},
  year={2020}
}

@inproceedings{radford2021clip,
  title={Learning Transferable Visual Models From Natural Language Supervision},
  author={A. Radford and J. W. Kim and C. Hallacy and A. Ramesh and G. Goh and S. Agarwal and G. Sastry and A. Askell and P. Mishkin and J. Clark and G. Krueger and I. Sutskever},
  booktitle={Proc. Int. Conf. Mach. Learn.},
  pages={8748--8763},
  year={2021}
}

@article{zhai2023siglip,
  title={Sigmoid Loss for Language Image Pre-Training},
  author={X. Zhai and B. Mustafa and A. Kolesnikov and L. Beyer},
  journal={arXiv preprint arXiv:2303.15343},
  year={2023}
}

@inproceedings{dosovitskiy2021vit,
  title={An Image is Worth 16x16 Words: Transformers for Image Recognition at Scale},
  author={A. Dosovitskiy and L. Beyer and A. Kolesnikov and D. Weissenborn and X. Zhai and T. Unterthiner and M. Dehghani and M. Minderer and G. Heigold and S. Gelly and J. Uszkoreit and N. Houlsby},
  booktitle={Proc. Int. Conf. Learn. Represent.},
  year={2021}
}

@article{hinton2015distilling,
  title={Distilling the Knowledge in a Neural Network},
  author={G. Hinton and O. Vinyals and J. Dean},
  journal={arXiv preprint arXiv:1503.02531},
  year={2015}
}

@inproceedings{zhang2019byot,
  title={Be Your Own Teacher: Improve the Performance of Convolutional Neural Networks via Self Distillation},
  author={L. Zhang and J. Song and A. Gao and J. Chen and C. Bao and K. Ma},
  booktitle={Proc. Int. Conf. Comput. Vis.},
  year={2019}
}

@article{hendrycks2019benchmarking,
  title={Benchmarking Neural Network Robustness to Common Corruptions and Perturbations},
  author={D. Hendrycks and T. Dietterich},
  journal={arXiv preprint arXiv:1903.12261},
  year={2019}
}

@article{hendrycks2020augmix,
  title={AugMix: A Simple Data Processing Method to Improve Robustness and Uncertainty},
  author={D. Hendrycks and N. Mu and E. D. Cubuk and B. Zoph and J. Gilmer and B. Lakshminarayanan},
  journal={arXiv preprint arXiv:1912.02781},
  year={2020}
}

@article{wong2025fontguard,
  title={{FontGuard}: A Robust Font Watermarking Approach Leveraging Deep Font Knowledge},
  author={K. Wong and J. Zhou and K. Li and Y. Si and X. Wu and J. Zhou},
  journal={IEEE Trans. Multimedia},
  year={2025},
  volume={27},
  pages={7876-7890},
  publisher={IEEE}
}

@inproceedings{wu2026editprint,
  title={Editprint: General Digital Image Forensics via Editing Fingerprint with Self-Augmentation Training},
  author={H. Wu and K. Li and Y. Li and J. Zhou},
  booktitle={IEEE Conf. Comput. Vis. Pattern Recog.},
  note={To appear},
  year={2026}
}

@inproceedings{li2026llm,
  title={{LLM} unlearning with {LLM} beliefs},
  author={K. Li and Q. Wang and Y. Wang and F. Li and J. Liu and B. Han and J. Zhou},
  booktitle={Proc. Int. Conf. Learn. Representat.},
  year={2026}
}
}

\end{document}